\def\paperID{XXXX}
\def\confName{CVPR}
\def\confYear{2026}

\def\paperTitle{LLaViDA: A Large Language Vision Driving Assistant for Explicit Reasoning and Enhanced Trajectory Planning}

\def\authorBlock{
    Yudong Liu$^{1}$\qquad
    Spencer Hallyburton$^{1}$\qquad
    Jiwoo Kim$^{1}$ \qquad
    Yueqian Lin$^{1}$ \qquad
    Yiming Li$^{1}$ \qquad
    Qinsi Wang$^{1}$ \\
    Hui Ye$^{2}$ \qquad
    Jingwei Sun$^{3}$ \qquad
    Miroslav Pajic$^{1}$ \qquad
    Yiran Chen$^{1}$ \qquad
    Hai Li$^{1}$ \\
    $^{1}$Duke University \qquad $^{2}$Georgia State University \qquad $^{3}$University of Florida \\
    {\tt\small $^{1}$\{first name.last name\}@duke.edu}
}

\newif\ifreview 
\newif\ifarxiv \newcommand{\arxiv}{\arxivtrue}
\newif\ifcamera 
\newif\ifrebuttal 

\arxiv 

\pdfoutput=1
\documentclass[10pt,twocolumn,letterpaper]{article}
\ifreview \usepackage[review]{cvpr} \fi
\ifarxiv \usepackage[pagenumbers]{cvpr} \fi
\ifrebuttal \usepackage[rebuttal]{cvpr} \fi
\ifcamera \usepackage{cvpr} \fi

\usepackage[utf8]{inputenc}
\usepackage{graphicx}	
\usepackage{amsmath}	
\usepackage{amssymb}	
\usepackage{booktabs}
\usepackage{times}
\usepackage{microtype}
\usepackage{epsfig}
\usepackage{caption}
\usepackage{float}
\usepackage{color, colortbl}
\usepackage{stfloats}
\usepackage{enumitem}
\usepackage{tabularx}
\usepackage{xstring}
\usepackage{multirow}
\usepackage{xspace}
\usepackage{url}
\usepackage{subcaption}
\usepackage{xcolor}
\usepackage[hang,flushmargin]{footmisc}
\usepackage{booktabs, multirow}
\usepackage{pifont}
\usepackage{listings}
\usepackage{xcolor}
\usepackage[section]{placeins}
\lstdefinestyle{tightcode}{
  basicstyle=\ttfamily\scriptsize,
  columns=fullflexible,
  breaklines=true,
  frame=single,
  numbersep=4pt,
  aboveskip=4pt, belowskip=4pt,
  showstringspaces=false,
}
\newcommand{\cmark}{\ding{51}}
\newcommand{\xmark}{\ding{55}}
\newcommand{\modelname}[1]{#1}

\ifcamera \usepackage[accsupp]{axessibility} \fi

% ...

\ifarxiv  \fi

\newcommand{\R}[1]{{%
    \textbf{%
        \ifstrequal{#1}{1}{\textcolor{red}{R#1}}{%
        \ifstrequal{#1}{2}{\textcolor{blue}{R#1}}{%
        \ifstrequal{#1}{3}{\textcolor{magenta}{R#1}}{%
        \ifstrequal{#1}{4}{\textcolor{teal}{R#1}}{%
                           \textcolor{cyan}{R#1}%
        }}}}%
    }%
}}

\usepackage{xr-hyper}

\makeatletter
\newcommand*{\addFileDependency}[1]{
  \typeout{(#1)}
  \@addtofilelist{#1}
  \IfFileExists{#1}{}{\typeout{No file #1.}}
}

\makeatother
\newcommand*{\myexternaldocument}[1]{
    \externaldocument{#1}
    \addFileDependency{#1.tex}
    \addFileDependency{#1.aux}
}

\definecolor{cvprblue}{rgb}{0.21,0.49,0.74}
\usepackage[pagebackref,breaklinks,colorlinks,allcolors=cvprblue]{hyperref}
\usepackage[capitalize]{cleveref}
\crefname{section}{Sec.}{Secs.}
\crefname{table}{Table}{Tables}
\crefname{figure}{Fig.}{Figs.}

\ifarxiv \crefname{appendix}{App.}{Apps.}
\else \crefname{appendix}{Suppl.}{Suppls.} \fi

\unless\ifarxiv \myexternaldocument{_supplementary} \fi

\begin{document}
\title{\paperTitle}
\author{\authorBlock}
\maketitle

\begin{abstract}
Trajectory planning is a fundamental yet challenging component of autonomous driving. End-to-end planners frequently falter under adverse weather, unpredictable human behavior, or complex road layouts, primarily because they lack strong generalization or few-shot capabilities beyond their training data. We propose \textbf{\modelname{LLaViDA}}, a \textbf{L}arge \textbf{La}nguage \textbf{Vi}sion \textbf{D}riving \textbf{A}ssistant that leverages a Vision-Language Model (VLM) for object motion prediction, semantic grounding, and chain-of-thought reasoning for trajectory planning in autonomous driving. A two-stage training pipeline—supervised fine-tuning followed by Trajectory Preference Optimization (TPO)—enhances scene understanding and trajectory planning by injecting regression-based supervision, produces a powerful ``VLM Trajectory Planner for Autonomous Driving.'' On the NuScenes benchmark, \modelname{LLaViDA} surpasses state-of-the-art end-to-end and other recent VLM/LLM-based baselines in open-loop trajectory planning task, achieving an average $\ell_{2}$ trajectory error of $\mathbf{0.31}\,\mathrm{m}$ and a collision rate of $\mathbf{0.10}\%$ on the \textit{NuScenes} test set. The code for this paper is available at \href{https://github.com/1999Lyd/LLaViDA} {\texttt{GitHub}}. 
\end{abstract}

\section{Introduction}
\label{sec:intro}
\begin{figure}[tp]
    \centering
    \includegraphics[width=\linewidth]{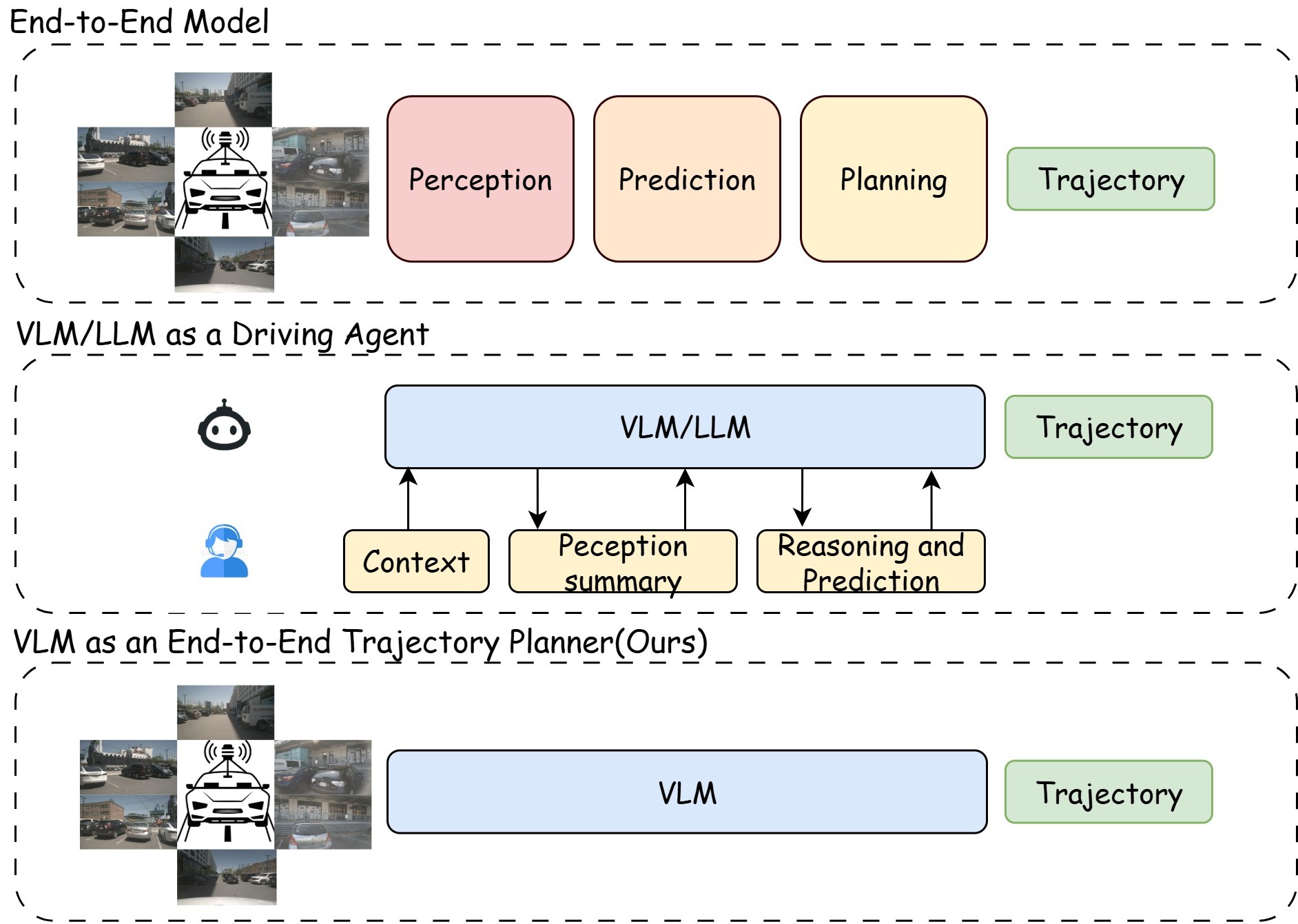}
    \caption{Three paradigms of tackling trajectory planning task in end-to-end autonomous driving.}
    \label{fig:data_sample}
   \vspace{-1mm}
\end{figure}
\begin{figure*}[t]
    \centering
    \includegraphics[width=0.9\linewidth]{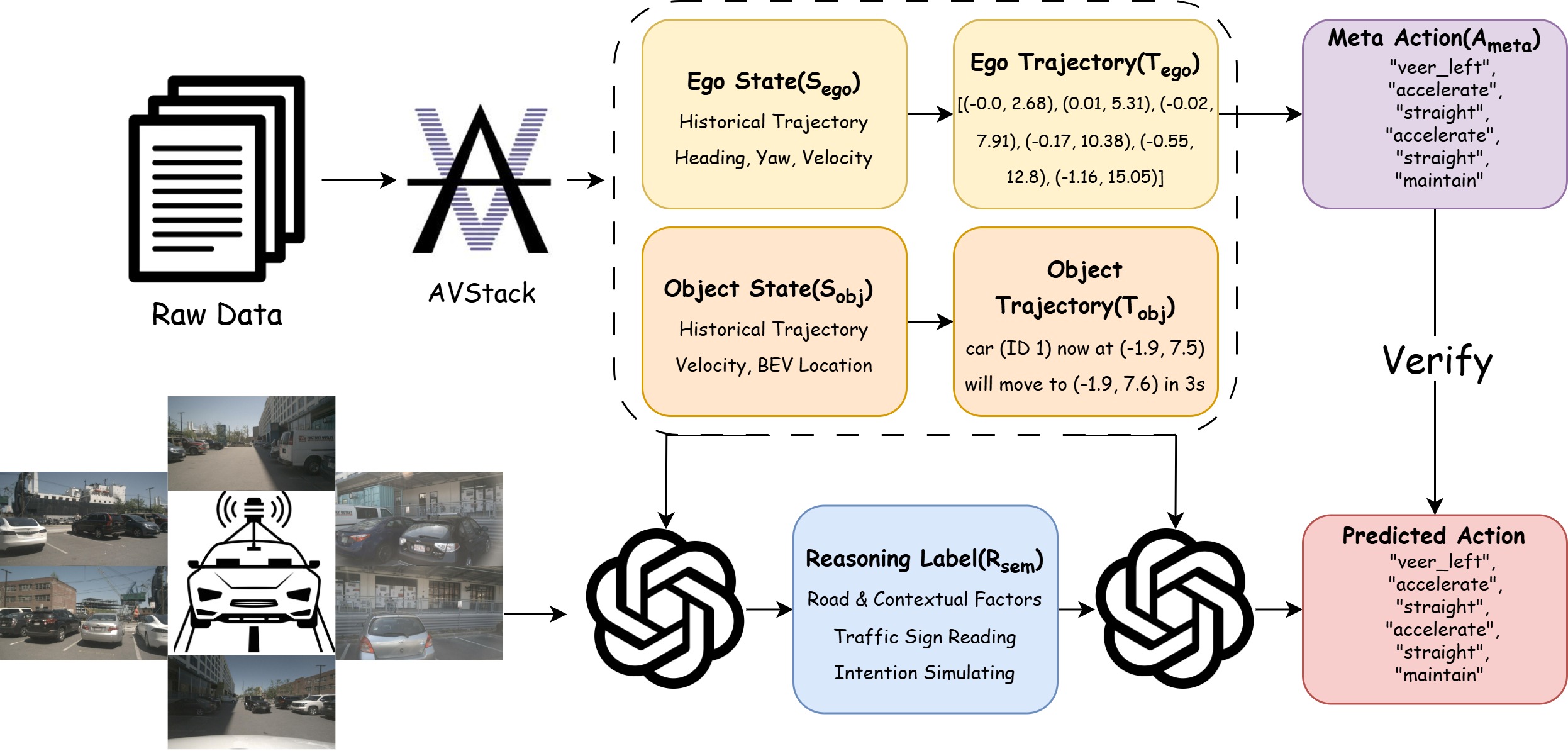}
    \caption{Construction pipeline of the proposed NuScenes-TP dataset. Starting from the raw NuScenes data, we extract ego and object states, derive their corresponding future trajectories, and further compute ego meta-actions from the ego trajectory. In parallel, GPT-4o is used to generate reasoning annotations, which are then validated against the ground-truth meta-actions.}
    \label{fig:data}
   
\end{figure*}
Trajectory planning transforms the dynamic visual environment into a safe and comfortable motion plan for autonomous vehicles. Conventional end-to-end models decompose this task into sequential modules—object detection, motion forecasting, occupancy prediction, and trajectory generation~\cite{hu2022stp3,hu2023_uniad}. End-to-end trajectory planners lack semantic understanding and exhibit limited few-shot generalization, which makes them prone to errors such as failing to follow traffic signs and struggling under uncommon conditions like adverse weather, atypical road layouts, or non-standard human behaviors.

Vision language models (VLMs) have recently demonstrated striking few-shot learning, semantic grounding, and chain-of-thought reasoning across heterogeneous vision language tasks, hinting at a unified alternative: recast perception, prediction, and planning as a single vision-language-conditioned reasoning problem~\cite{guo2024vlm,xu2024vlm,guo2025vdt,you2024v2x,zhou2025opendrivevla}. However, two obstacles stand in the way. First, without specialized training, general-purpose VLMs struggle to generate structured and numerically precise trajectory plans within a single inference step. Instead, they require multi-turn interactions to iteratively refine the output into a valid trajectory format ~\cite{mao2023language,mao2023gpt}, which introduces substantial and impractical latency that hinders real-world deployment. Second, most existing autonomous driving datasets lack structured natural-language rationales or action traces that explicitly connect scene understanding to the corresponding ground-truth trajectory—information that would teach a VLM how to reason through the scene and derive a proper driving plan, depriving the model of essential supervision~\cite{caesar2020nuscenes,sun2020scalability,huang2018apolloscape,mao2021one,ma2024lampilot}.

We address these challenges with \textbf{\modelname{LLaViDA}}, a \textbf{L}arge \textbf{La}nguage \textbf{Vi}sion \textbf{D}riving \textbf{A}ssistant for trajectory planning. Only taking camera images as input, the model produces a \emph{hierarchical chain-of-thought} that forecasts the motion of all salient traffic participants, describes scene semantics together with road layout and weather conditions, infers ego-vehicle intentions in context, derives a suitable meta-action (e.g., lane change, maintain speed), and finally emits a numerically precise low-risk trajectory that respects vehicle dynamics and traffic rules. Training proceeds in two stages. First, supervised vision–language fine-tuning grounds visual tokens in traffic semantics. Second, \emph{Trajectory Preference Optimization} (TPO) further optimizes trajectory quality by injecting regression-based supervision without requiring additional annotations. For each training prompt, we sample multiple complete outputs from the checkpoint after supervised fine-tuning and score their trajectories by the \(\ell_{2}\) distance to the ground-truth path; these scores define preferred vs.\ dispreferred pairs for TPO. This replaces purely token-level supervision with a continuous, trajectory-quality signal—injecting a regression-like supervision into the generative objective—so the VLM learns to discriminate subtle geometric differences between candidate paths and consistently prefer lower-error plans.

To enable this process, we curate \textbf{NuScenes-TP}, a trajectory-planning dataset derived from the public NuScenes~\cite{caesar2020nuscenes} corpus and enriched with ground-truth meta action sequences defined by crafted rules and a natural-language reasoning process generated by GPT-4o. The empirical results confirm the effectiveness of our approach. On the NuScenes evaluation benchmark, \modelname{LLaViDA} substantially reduces the average displacement error and the collision rate, outperforming both End-to-End planning pipelines and contemporary VLM/LLM-based baselines. These findings demonstrate that language-conditioned reasoning, enhanced by reinforcement learning, yields a robust and deployable VLM-based trajectory-planning system.

\vspace{0.5\baselineskip}
In summary, our contributions are:
\begin{itemize}
\item We introduce \textbf{\modelname{LLaViDA}}, an open-source framework that converts raw camera images into physically feasible low-risk trajectories with interpretable reasoning traces in a single turn of inference. \modelname{LLaViDA} achieves new state-of-the-art in open-loop trajectory planning.
\item We develop and release \textbf{NuScenes-TP}, a reasoning-augmented, VLM-compatible trajectory planning dataset that bridges conventional autonomous driving datasets and large-scale VLM training for trajectory planning task.
\item We propose a data efficient training pipeline that integrates regression supervision with Trajectory Preference Optimization. We train on only $23\text{k}$ curated NuScenes-TP samples, yet we adapt a generic VLM into a decent “Trajectory Planning Expert”.
\end{itemize}

\section{Related Work}
\label{sec:related}

\paragraph{Vision–Language Models.} 
Vision–Language Models (VLMs) have emerged as a central research topic in the computer vision community~\cite{bai2025qwen2,li2023blip,li2024llava,liu2023visual}. Pretrained on large-scale image–text pairs and subsequently fine-tuned with extensive vision instruction-tuning corpora, VLMs acquire strong commonsense reasoning capabilities, broad world knowledge, and robust in-context learning abilities. Building on these foundations, numerous works have explored their application to downstream specialized domains, such as robotics~\cite{zhao2025cot,shukor2025smolvla,intelligence2025pi_}, embodied AI~\cite{zhang2025embodied,luo2025visual}, and autonomous driving~\cite{guo2024vlm,xu2024vlm,guo2025vdt,you2024v2x,zhou2025opendrivevla}. These domain-specific systems are typically obtained by fine-tuning a base VLM on carefully curated datasets, transforming it into an expert for the target domain.

\paragraph{Traditional End-to-End Models for Trajectory Planning.}
Classical trajectory planning frameworks, such as UniAD~\cite{hu2023_uniad} and ST-P3~\cite{hu2022stp3}, decompose the problem into a sequence of subtasks: object detection, motion prediction, occupancy prediction, and trajectory generation. Each subtask is handled by a dedicated module, often implemented as a stack of transformer or convolutional blocks followed by a classification or regression head. Although such modules can be highly effective in their respective domains, they lack the generalized knowledge and zero- or few-shot abilities of large language or vision language models. Consequently, they struggle with novel scenes or unseen targets, leading to suboptimal output. Furthermore, since these models operate largely as black boxes without explicit reasoning traces, they offer limited interpretability, a key concern for safety-critical systems such as autonomous driving.

\paragraph{VLM/LLM for Trajectory Planning.}
Given their strong zero-shot generalization and reasoning abilities, VLMs and LLMs have recently been investigated for trajectory planning in autonomous driving. Some approaches distill the semantic understanding of VLMs into traditional end-to-end planning models to enhance perception and decision-making~\cite{liu2025vlm, feng2025verdi}. Others integrate VLMs alongside conventional modules in a dual-system design, where the VLM acts as a high-level planner that complements low-level perception and control~\cite{tian2024drivevlm}. A different line of work treats the VLM/LLM as an autonomous agent: by providing it with rich contextual information, such as ego-vehicle states, surrounding vehicle dynamics, road conditions, and historical cases stored in a memory base, through multi-turn interactions, the agent can produce an appropriate trajectory~\cite{mao2023gpt, mao2023language}. 

Our work differs from previous efforts by leveraging a single VLM as the sole decision-making engine for trajectory planning. Rather than acting as an auxiliary module or requiring multi-turn dialogues, our system directly generates state-of-the-art trajectory predictions, using only camera images as inputs and producing both the planned trajectory and an explicit reasoning process in \textbf{a single turn of inference}.

\section{Method}
\label{sec:method}

\begin{figure*}[t]
    \centering
    \includegraphics[width=\linewidth]{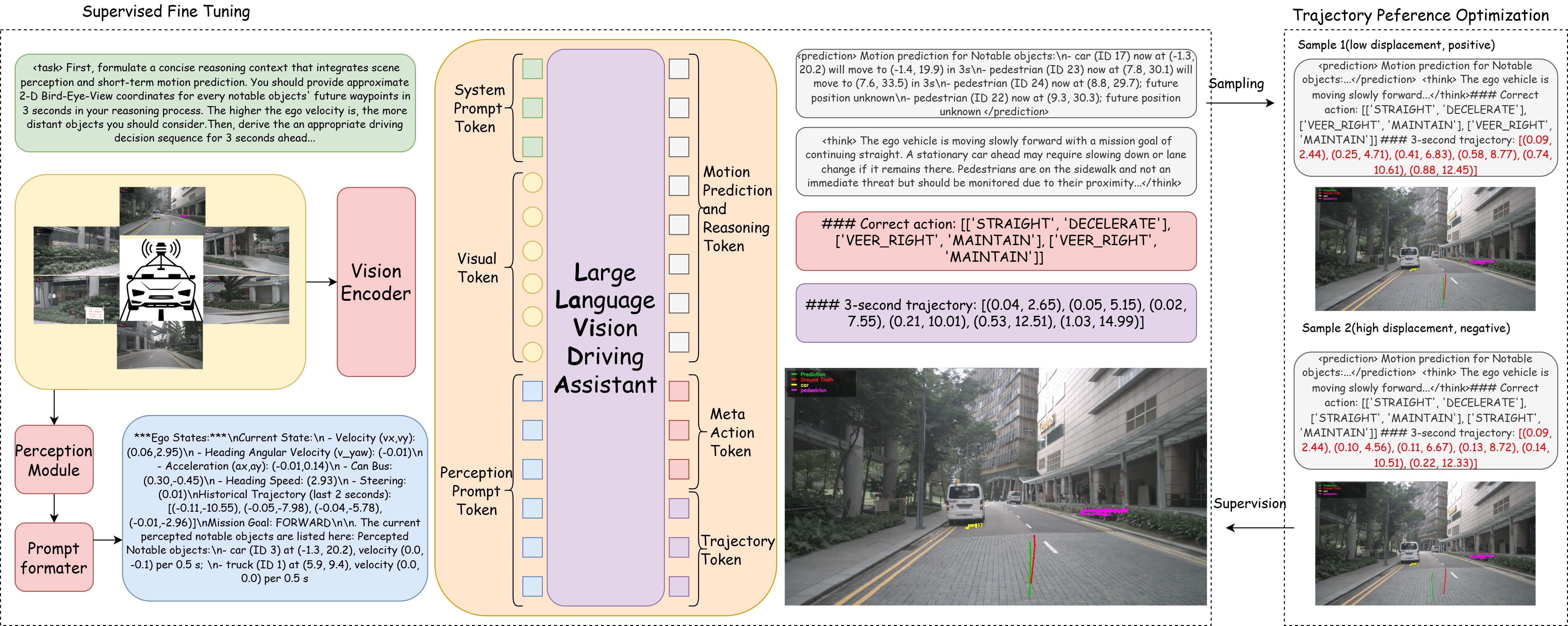}
    \caption{Overview of the proposed LLaViDA framework. LLaViDA models trajectory planning as a multi-object motion-prediction problem. By explicitly predicting the motion of key objects in the scene (yellow and purple traces) and imitating human driver reasoning, it generates an accurate ego trajectory (green) in a causally grounded manner.}

    \label{fig:pipeline}
    \vspace{-1mm}
\end{figure*}
\subsection{Preliminary}
\paragraph{Autoregressive Language Models.}
Let $\mathbf{x}=(x_{1},\dots,x_{N})$ denote a prompt of $N$ discrete tokens and $\mathbf{y}=(y_{1},\dots,y_{T})$ the target continuation of length $T$.  An autoregressive language model parameterized by $\theta$ defines a left-to-right factorisation
\begin{equation}
    p_{\theta}(\mathbf{y}\mid\mathbf{x}) \;=\;
    \prod_{t=1}^{T} p_{\theta}\!\bigl(y_{t}\,\bigl|\;\mathbf{x},\,y_{<t}\bigr),
    \label{eq:autoregressive}
\end{equation}
where $y_{<t}$ is the prefix $(y_{1},\dots,y_{t-1})$.  Training maximizes the logarithmic likelihood of observed sequences, \emph{i.e.} $\max_{\theta}\!\sum_{(\mathbf{x},\mathbf{y})}\!\log p_{\theta}(\mathbf{y}\mid\mathbf{x})$, and generation proceeds by iterative sampling or decoding from~\eqref{eq:autoregressive}.

\paragraph{Vision–Language Models.}
A Vision–Language Model augments the text-only backbone with a vision encoder $f_{v}$ ~\cite{zhang2022dino,radford2021learning,zhai2023sigmoid}and a projection head $W\!\in\!\mathbb{R}^{d\times d_{v}}$ that aligns visual features to the language embedding space of dimensionality~$d$.  An image $I$ is partitioned into patches and encoded to a sequence of $M$ \emph{vision tokens} $\mathbf{v}=f_{v}(I)\in\mathbb{R}^{M\times d_{v}}$; after linear projection $\tilde{\mathbf{v}}=W\mathbf{v}$ the combined input becomes
\[
   \mathbf{z}
   \;=\;
   (\underbrace{\langle\text{img}\rangle,\tilde{\mathbf{v}}_{1},\dots,\tilde{\mathbf{v}}_{M}}_{\text{visual stream}},
   \;\underbrace{\langle\text{txt}\rangle,x_{1},\dots,x_{N}}_{\text{text stream}}),
\]
which is fed to the same transformer decoder that realizes the language model.  The cross-modal attention learned during pretraining aligns image regions and textual concepts, enabling text generation conditioned on visual input.

\paragraph{Chain-of-Thought Reasoning.}
Beyond direct answers, large language models can emit an explicit \emph{chain of thought} (CoT) $\mathbf{r}=(r_{1},\dots,r_{K})$ that records intermediate reasoning steps~\cite{wei2022chain}.  In generative form, the model factors the joint distribution.
\begin{equation}
   p_{\theta}(\mathbf{r},\mathbf{y}\mid\mathbf{z})
   \;=\;
   \prod_{k=1}^{K} p_{\theta}\!\bigl(r_{k}\mid\mathbf{z},r_{<k}\bigr)
   \prod_{t=1}^{T} p_{\theta}\!\bigl(y_{t}\mid\mathbf{z},\mathbf{r},y_{<t}\bigr),
   \label{eq:cot}
\end{equation}
so that the final prediction $\mathbf{y}$ is \emph{conditioned} on the full reasoning trace $\mathbf{r}$.  In our trajectory-planning setting the CoT first enumerates motion predictions for salient objects, describes scene semantics (road geometry, weather, traffic signs), infers ego intentions, and proposes a high-level meta-action; the concluding token segment encodes a structured trajectory consistent with all prior steps.

\subsection{NuScenes-TP Construction}
\label{sec:nuscenes}
The raw NuScenes data lack structured natural-language annotations that can directly supervise reasoning or action understanding in VLMs.
Therefore, we transform raw NuScenes~\cite{caesar2020nuscenes} scenes into a dataset tailored for fine-tuning an action-driven VLM using the AVstack framework~\cite{hallyburton2023avstack}. Each training sample is a longitudinal sequence of 40 frames sampled at 2 Hz, yielding approximately 20 seconds of multimodal driving context per scene.

At each frame, the pipeline collects only information that would be available to the ego vehicle in real time to be used as input into the model:
\begin{itemize}
    \item ego state (pose, velocity, acceleration and mission goal);  
    \item ego trajectory history up to the current frame;
    \item ego 3-second future trajectory starting from the current frame;
    \item images from six calibrated surround-view cameras;  
    \item key objects within a fixed distance of the ego;  
    \item instantaneous key object states (class, position, velocity, yaw);  
    \item key objects' trajectory histories up to the current frame. 
    \item key objects' 3-second future trajectories starting from the current frame;
\end{itemize}
All objects are represented in a bird's-eye view (BEV) coordinate system. Past trajectories are registered across frames and expressed in an ego-local reference frame for temporal consistency.
The ground truth trajectory waypoints are generated by looking ahead in the CAN-bus data and projecting future ego states into BEV at 0.5\,s intervals up to 3\,s. In parallel, discrete meta actions are derived along \texttt{Lateral} and \texttt{Longitudinal} axes using AVstack action evaluators. Longitudinal actions capture high-level acceleration modes (\texttt{REVERSE}, \texttt{BRAKE TO STOP}, \texttt{DECELERATE}, \texttt{MAINTAIN}, \texttt{ACCELERATE}), while lateral actions describe maneuver intent (\texttt{TURN LEFT}, \texttt{CHANGE LANE LEFT}, \texttt{VEER LEFT}, \texttt{STRAIGHT}, \texttt{VEER RIGHT}, \texttt{CHANGE LANE RIGHT}, \texttt{TURN RIGHT}).  

This dual approach of yielding both continuous-space waypoints and discrete action labels enables both trajectory-prediction objectives and action-conditioned VLM training. In particular, the derived \textbf{Meta Action} label serves 
 as a high-level natural language abstraction, which bridges the gap between the reasoning process and the numeric trajectory labels, making it easier for the VLM to learn the correspondence between reasoning and the predicted trajectory.
 
 To introduce reasoning trace supervision in the VLM fine-tuning, we leverage \textsc{GPT-4o}~\cite{achiam2023gpt} to synthesize the reasoning label that describes weather/road conditions, scene semantics and driver intentions.
The corresponding ground-truth meta action provides a quantitative reference for validating these reasoning labels. Formally, we synthesize the reasoning label $r$ and the meta action $a$ as
\begin{align}
r &= \mathrm{GPT}(V, p),\\
a &= \mathrm{GPT}(V, r, p),
\end{align}
where $V$ denotes visual input and $p$ denotes perception/context and instruction prompt.
We accept $r$ as a validated reasoning label iff $a = a^{\star}$, where $a^{\star}$ is the ground-truth meta action. This procedure enforces the quality of the reasoning labels with minimum additional annotations.

The resulting dataset is serialized in JSON format, with each entry containing metadata (action tables, versioning), per-frame ego state, six-camera image paths with calibration, BEV waypoints, discrete meta-actions, object trajectories and synthesized reasoning process. Each official split (\texttt{train}, \texttt{test}) is stored separately for compatibility with the subsequent training and evaluation.

\subsection{Supervised Fine-Tuning (SFT)}
\label{sec:sft}

We initiate our supervised fine-tuning from a general foundation VLM. Following~\cite{liu2024llavanext}, each input image is uniformly partitioned into four tiles, with each tile encoded into the same number of vision tokens as the original image, thereby enhancing sensitivity to small or distant objects. While this tiling increases computational and memory costs, we mitigate the overhead by applying a $2{\times}2$ average pooling operation on the vision token grid before feeding it to the language model, following \citep{zhang2024llavanext-video,li2025tokenpacker}, which demonstrate that such post-training pooling preserves most of the visual details across VLM tasks.

For trajectory planning, the visual input comprises six camera views $\{I^{(k)}\}_{k=1}^{6}$. Let $E_{\mathrm{v}}$ produce per-tile tokens $V^{(k)}\in\mathbb{R}^{H\times W\times d}$. We form
\begin{equation}
    \bar V^{(k)} = \mathrm{AvgPool}_{2\times 2}\!\big(V^{(k)}\big), \quad
    \bar V = \mathrm{Concat}_{k=1}^{6} \bar V^{(k)},
\end{equation}
where $\bar V$ is the pooled visual prefix supplied to the VLM.

Because NuScenes-TP provides basic states per object (BEV location, velocity, and class), we use these directly during VLM training for simplicity, which correspond to the perception prompt part in ~\ref{fig:pipeline}. At evaluation time, we instead run a lightweight 3D detector to estimate the same states following ~\cite{zhou2025opendrivevla}. We adopt BEVFormer~\cite{li2024bevformer} as it relies only on camera inputs. BEVFormer extracts each object's absolute translation and velocity within the map, and the absolute translation is then projected onto 2D BEV coordinates relative to the ego vehicle. From the detections, we further select the critical object set $\mathcal{C}$ within a radius $L(v_{\mathrm{ego}})$ that adapts to the speed of the ego vehicle,

\begin{equation}
    L(v_{\mathrm{ego}}) = L_{0} + \kappa\|v_{\mathrm{ego}}\|,
\end{equation}
where $L_0$ and $\kappa\ $ are predefined hyperparameters. We serialize their states $s_i = <p_i, v_i, \mathrm{cls}_i>$ in structured natural language.

In conclusion, the SFT input $I$ combines: (1) ego state $S_{\mathrm{ego}}$ (velocity, acceleration, yaw, 2\,s trajectory history and mission goal), (2) critical-object states $S_{\mathrm{obj}} = \{s_i\}_{i\in\mathcal{C}}$, and (3) a schema prompt $p$ guiding structured output:
\begin{equation}
    I = <S_{\mathrm{ego}}, S_{\mathrm{obj}}, p>.
\end{equation}
The supervised target $O$ contains a reasoning trace and actionable outputs:
\begin{equation}
    O = <\langle\texttt{think}\rangle
        T_{\mathrm{obj}}, R_{\mathrm{sem}}
        \langle/\texttt{think}\rangle,\;
        A_{\mathrm{meta}},\;
        T_{\mathrm{ego}}>.
\end{equation}
Here, $T_{\mathrm{obj}}$ are 3\,s future BEV trajectories of all critical objects; $R_{\mathrm{sem}}$ encodes the semantics of the scene, the road/weather conditions, and the interpretation of the intent of the driver; $A_{\mathrm{meta}}$ is the high-level meta-action; and $T_{\mathrm{ego}}=\{({w_1}^{\ast}_t,{w_2}^{\ast}_t)\}_{t=1}^3$ is the ground truth 3\,s ego trajectory, where $w_1$ and $w_2$ correspond to the lateral axis and the longitudinal axis respectively.

Let $p_{\theta}(\cdot\,|\,\bar V, I)$ be the token probability of the VLM.  
The SFT loss is as follows:
\begin{align}
    \mathcal{L}_{\mathrm{SFT}}(\theta)
    &= -\sum_{t\in\mathcal{I}_{\mathrm{all}}}
       w_t\,\log p_{\theta}\big(o_t \mid \bar V, I, o_{<t}\big) \label{eq:sft-loss} \\
    &= \mathcal{L}_{\mathrm{reason}}
     + \lambda\,(\mathcal{L}_{\mathrm{traj}}
     + \mathcal{L}_{\mathrm{meta}}), \nonumber
\end{align}
where $\mathcal{L}_{\mathrm{reason}}$ covers reasoning tokens inside \texttt{<think>...</think>}, $\mathcal{L}_{\mathrm{traj}}$ covers numeric trajectory tokens and $\mathcal{L}_{\mathrm{meta}}$ covers meta-action tokens, with weight $\lambda\ge 0$. Empirically, we found that the setting $\lambda = 1.2$ yields the best results.

\subsection{Trajectory Preference Optimization (TPO)}
\label{sec:dpo}

\noindent\textbf{Motivation.}
Cross-entropy in SFT optimizes token classification in a discrete space, whereas trajectory prediction is inherently continuous. As a result, with the token-level cross-entropy of SFT, the model primarily maximizes the likelihood of the ground-truth trajectory while collapsing all alternatives into a single 'wrong' class - assigning nearly the same penalty to both small and large geometric deviations, offering little incentive to prefer numerically closer plans. To inject a regression-like signal without collecting additional annotations or introducing structural change to the model, we adopt reinforcement learning via \emph{Trajectory Preference Optimization} (TPO), a downstream application of Direct Preference Optimization that uses preference pairs instead of an explicit reward model~\cite{rafailov2023direct}.

\noindent\textbf{Pair construction.}
We initialize the policy \(\pi_{\theta}\) and a frozen reference \(\pi_{\mathrm{ref}}\) from the best SFT checkpoint. For each training instance \((\bar V, I)\), we sample \(K=16\) complete responses \(y^{(k)}\) from \(\pi_{\mathrm{ref}}\) with temperature \(\tau=1.5\). Each response contains the chain of thought and a numeric ego trajectory \(T_{\mathrm{ego}}\). We score the response \(k\) based on the average \(\ell_{2}\) displacement between the sampled way-points $ \hat{w} $ and the ground-truth way-points ${w}^{\ast} $:
\begin{align}
d_{k}
= \frac{1}{3}\sum_{t=1}^{3}
\big\|(\hat{w_1}^{(k)}_{t},\hat{w_2}^{(k)}_{t})-({w_1}^{\ast}_{t},{w_2}^{\ast}_{t})\big\|_{2}.
\end{align}
We then set \(y^{+}=\arg\min_{k} d_{k}\) and \(y^{-}=\arg\max_{k} d_{k}\), forming
\(\mathcal{D}_{\mathrm{TPO}}=\{(\bar V, I, y^{+}, y^{-})\}\).

\noindent\textbf{Objective.}
Let \(\Delta_{\theta}=\log\pi_{\theta}(y^{+}\mid \bar V,I)-\log\pi_{\theta}(y^{-}\mid \bar V,I)\) and
\(\Delta_{\mathrm{ref}}=\log\pi_{\mathrm{ref}}(y^{+}\mid \bar V,I)-\log\pi_{\mathrm{ref}}(y^{-}\mid \bar V,I)\).
With the logistic function \(\sigma(u)=1/(1+e^{-u})\) and scale \(\beta=0.1\), the TPO loss is
\begin{align}
\mathcal{L}_{\mathrm{TPO}}(\theta)
= -\,\mathbb{E}_{(\bar V,I,y^{+},y^{-})\sim \mathcal{D}_{\mathrm{TPO}}}
\Big[\log \sigma\big(\beta(\Delta_{\theta}-\Delta_{\mathrm{ref}})\big)\Big].
\label{eq:dpo}
\end{align}

\noindent\textbf{Effect.}
TPO introduces a displacement-aware \emph{regression} signal on top of token likelihoods. By ranking sampled trajectories with the continuous \(\ell_{2}\) distance and selecting preferred/dispreferred pairs, the objective explicitly rewards lower-displacement plans and penalizes higher-displacement ones; because pairs are drawn from a continuum of \(d_k\) values, the effective penalty varies with the displacement gap rather than treating all errors equally. This displacement-conditioned training signal yields smoother supervision than pure cross-entropy and strengthens the link between textual reasoning and geometric accuracy.

\section{Experiments}
\subsection{Experimental Setup}
\label{sec:exp-setup}

\paragraph{Dataset.}
All experiments are conducted on the NuScenes, a widely used autonomous-driving benchmark. We follow the standard split of NuScenes to obtain a training set of \textbf{23,423} samples(NuScenes-TP) and a test set of \textbf{6,019} samples. Supervised fine-tuning (SFT) is performed on the 23k training samples. For TPO, we use the best SFT checkpoint as the reference policy, sample $K{=}16$ full outputs per training instance at temperature $\tau{=}1.5$, compute the $\ell_2$ displacement of each sampled trajectory to the ground truth over a 3\,s horizon, and form preference pairs by selecting the minimum displacement sample as positive and the maximum displacement sample as negative. All main evaluations are reported on the test split.

\paragraph{Baselines.}
We compare against recent planning systems spanning both \emph{modular / non-autoregressive} pipelines and \emph{autoregressive (VLM/LLM)} approaches, including strong implementations representative of state-of-the-art.

\paragraph{Models.}
We select LLaVA-NeXT-LLaMA3-8B~\cite{li2024llava} as our main foundation VLM. We also performed additional ablation studies on Qwen2.5-VL-7B~\cite{bai2023qwen} and InternVL-3.5-8B~\cite{chen2024internvl}.

\paragraph{Metrics.}
Since our focus is trajectory planning, we adopt two standard metrics measured over a 3\,s horizon: (i) average $\ell_2$ displacement error between the predicted ego trajectory and ground truth ego trajectory(the results aggregations are different under 2 protocols, therefore yielding different results; details in ~\ref{app:metrics}),
\begin{equation}
\ell_2 \;=\; \tfrac{1}{H}\sum_{t=1}^{H} \lVert (\hat{w_1}_t,\hat{w_2}_t) - ({w_1}_t^\ast,{w_2}_t^\ast) \rVert_2,
\end{equation}
and (ii) \emph{collision rate}, the fraction of samples for which the planned ego footprint intersects any annotated obstacle within the horizon. Because two evaluation protocols are prevalent in the literature, we report results under both the \textsc{ST-P3} and \textsc{UniAD} settings to enable comprehensive and fair comparisons (detailed metrics calculation in ~\ref{app:metrics}).

\subsection{Main Results}
\label{sec:main-results}

We compare our method with recent trajectory planners \emph{non-autoregressive} (modular) and \emph{autoregressive} (VLM / LLM) under both prevalent evaluation protocols. For open-source baselines, we verified metric implementations and, when result under only one protocol was provided, we re-evaluated their released trajectories under the other protocol for completeness. For baselines without code, we report the numbers from their papers; unless explicitly stated otherwise, we treat their protocol as \textsc{ST-P3} by default.

\begin{table*}[t]
\centering
\setlength{\tabcolsep}{2.8pt}
\renewcommand{\arraystretch}{1.05}
\scriptsize
\caption{Comparison on NuScenes under the two standard protocols. We report per-horizon and averaged $\ell_2$ displacement (m, $\downarrow$) and collision rate (\%, $\downarrow$). Dashes indicate metrics not reported and not reproducible from available papers.}
\label{tab:main}
\resizebox{\textwidth}{!}{\begin{tabular}{l
                cccc cccc
                cccc cccc
                l}
\toprule
& \multicolumn{8}{c}{\textbf{ST-P3 protocol}} & \multicolumn{8}{c}{\textbf{UniAD protocol}} & \\
\cmidrule(lr){2-9}\cmidrule(lr){10-17}
\multirow{2}{*}{\textbf{Method}} &
\multicolumn{4}{c}{$\ell_2$ (m) $\downarrow$} & \multicolumn{4}{c}{Collision (\%) $\downarrow$} &
\multicolumn{4}{c}{$\ell_2$ (m) $\downarrow$} & \multicolumn{4}{c}{Collision (\%) $\downarrow$} &
\multirow{2}{*}{\textbf{Backbone LLM/VLM}} \\
\cmidrule(lr){2-5}\cmidrule(lr){6-9}\cmidrule(lr){10-13}\cmidrule(lr){14-17}
& 1s & 2s & 3s & Avg. & 1s & 2s & 3s & Avg. & 1s & 2s & 3s & Avg. & 1s & 2s & 3s & Avg. & \\
\midrule
\multicolumn{18}{l}{\textit{Non-autoregressive methods}} \\
ST-P3 \citep{hu2022stp3}          & 1.33 & 2.11 & 2.90 & 2.11 & 0.23 & 0.62 & 1.27 & 0.71 & 1.51 & 2.85 & 4.22 & 2.86 & 0.18 & 0.78 & 1.96 & 0.97 & — \\
VAD \citep{jiang2023vad}             & 0.17 & 0.34 & 0.60 & 0.37 & 0.07 & 0.10 & 0.24 & 0.14 & 0.23 & 0.66 & 1.31 & 0.73 & 0.04 & 0.21 & 0.55 & 0.27 & — \\
UniAD \citep{hu2023_uniad}         & 0.44 & 0.67 & 0.96 & 0.69 & 0.04 & 0.08 & 0.23 & 0.12 & 0.48 & 0.96 & 1.65 & 1.03 & 0.05 & 0.17 & 0.71 & 0.31 & — \\
InsightDrive \citep{insightdrive2025}
                            & 0.23 & 0.41 & 0.68 & 0.44 & 0.09 & 0.10 & 0.27 & 0.15 & 0.30 & 0.72 & 1.41 & 0.81 & 0.08 & 0.15 & 0.84 & 0.36 & — \\
\midrule
\multicolumn{18}{l}{\textit{Autoregressive methods}} \\
DriveVLM \citep{tian2024drivevlm}   & 0.18 & 0.34 & 0.68 & 0.40 & 0.10 & 0.22 & 0.45 & 0.27 & — & — & — & — & — & — & — & — & Qwen-VL-7B \citep{bai2023qwen} \\
GPT-Driver \citep{mao2023gpt} & 0.20 & 0.40 & 0.70 & 0.44 & 0.04 & 0.12 & 0.36 & 0.17 & 0.27 & 0.74 & 1.52 & 0.84 & 0.07 & 0.15 & 1.10 & 0.44 & GPT-3.5 \\
RDA-Driver \citep{huang2024making} & 0.17 & 0.37 & 0.69 & 0.40 & \textbf{0.01} & 0.05 & 0.26 & \underline{0.10} & 0.23 & 0.73 & 1.54 & 0.80 & \textbf{0.00} & 0.13 & 0.83 & 0.32 & LLaVA-7B \\
Agent-Driver \citep{mao2023language}
                            & 0.16 & 0.34 & 0.61 & 0.37 & \underline{0.02} & \textbf{0.07} & \textbf{0.18} & \textbf{0.09} & 0.22 & 0.65 & 1.34 & 0.74 & \underline{0.02} & 0.13 & \textbf{0.48} & \textbf{0.21} & GPT-3.5-Turbo \\
OpenDriveVLA \citep{zhou2025opendrivevla}
                            & 0.15 & 0.31 & 0.55 & 0.33 & \textbf{0.01} & 0.08 & 0.21 & \underline{0.10} & \underline{0.20} & \underline{0.58} & \underline{1.21} & \underline{0.66} & \textbf{0.00} & 0.22 & 0.55 & 0.25 & Qwen2.5-7B \citep{bai2025qwen2} \\
EMMA \citep{hwang2024emma}           & \underline{0.14} & \underline{0.29} & \underline{0.54} & \underline{0.32} & — & — & — & — & — & — & — & — & — & — & — & — & Gemini \citep{team2024gemini} \\
OpenEMMA \citep{openemma}   & 1.45 & 3.21 & 3.76 & 2.81 & 0.31 & 0.77 & 1.45 & 0.84 & 1.67 & 4.11 & 5.12 & 3.63 & 0.32 & 0.89 & 1.64 & 0.95 & Qwen-VL-7B \citep{bai2023qwen} \\
\textbf{LLaViDA (ours)}     & \textbf{0.14} & \textbf{0.28} & \textbf{0.51} & \textbf{0.31} &
                              0.03 & \textbf{0.07} & \underline{0.19} & \underline{0.10} &
                              \textbf{0.19} & \textbf{0.54} & \textbf{1.09} & \textbf{0.61} &
                              0.06 & \textbf{0.09} & \underline{0.50} & \underline{0.22} & LLaMA-3-8B \\
\bottomrule
\end{tabular}}
\vspace{-5mm}
\end{table*}

As shown in Table~\ref{tab:main}, \textbf{LLaViDA} achieves state-of-the-art performance among recent methods. Notably, our model surpasses both \emph{Agent-Driver} and \emph{EMMA} on $\ell_2$ displacement error, despite those methods leveraging more powerful \emph{closed-source} LLM backbones (GPT-3.5 and Gemini, respectively) compared to our open-source LLaMA-3-8B foundation. We attribute this performance gain to our \emph{trajectory preference optimization} approach: the TPO stage (Sec.~\ref{sec:dpo}) incorporates an explicit trajectory-quality signal—$\ell_2$ displacement from ground truth over a 3\,s planning horizon—to introduce regression-based supervision into VLM trajectory planning. By consistently favoring low-error trajectory generations during training, TPO demonstrates a promising direction for improving the trajectory-planning capabilities of vision-language models.

\subsection{Ablation Studies}
\label{sec:ablation}

\noindent\textbf{Effect of training sample components.}
Our \textit{NuScenes-TP} supervision for SFT comprises (input side) camera images, ego state (2\,s history, velocity, mission goal) and \emph{key-object} states (BEV location, velocity, class), and (label side) key-object motion forecasts, natural-language reasoning, and a discrete meta-action preceding the numeric trajectory. We remove one component at a time and re-train to quantify its contribution. Results in Table~\ref{tab:ablate} show that \emph{textual} input—especially the \emph{key-object state}—is most critical to accurate planning; our hypothesis is that purely visual conditioning can induce VLM hallucination under challenging scenes, a trend corroborated by the case studies in Section~\ref{sec:qualitative}. On the label side, \emph{meta-action} has the largest impact, indicating it effectively bridges textual reasoning and precise trajectory generation. Forecasting future motion of key objects mitigates prospective occlusions within the 3\,s horizon and further improves accuracy.


\begin{table}[t]
\centering
\setlength{\tabcolsep}{3pt}
\renewcommand{\arraystretch}{1.05}
\footnotesize
\caption{Ablation on \textit{NuScenes-TP} components. We report average $\ell_2$ (m, $\downarrow$) and collision rate (\%, $\downarrow$) over 3\,s under both \textsc{UniAD} and \textsc{ST-P3} protocols.}
\label{tab:ablate}
\begin{tabular}{ccccc|cc|cc}
\toprule
\multicolumn{5}{c}{\textbf{Components}} & \multicolumn{2}{c}{\textbf{$\ell_2$ (m) $\downarrow$}} & \multicolumn{2}{c}{\textbf{Collision (\%) $\downarrow$}} \\
\cmidrule(lr){1-5}\cmidrule(lr){6-7}\cmidrule(lr){8-9}
\multicolumn{2}{c}{\textit{Input}} & \multicolumn{3}{c}{\textit{Label}} & \multirow{2}{*}{UniAD} & \multirow{2}{*}{ST\text{-}P3} & \multirow{2}{*}{UniAD} & \multirow{2}{*}{ST\text{-}P3} \\
\cmidrule(lr){1-2}\cmidrule(lr){3-5}
Ego & Obj & Motion & Reas. & Meta & & & & \\
\midrule
\xmark & \cmark & \cmark & \cmark & \cmark & 0.89 & 0.48 & 0.38 & 0.17 \\
\cmark & \xmark & \cmark & \cmark & \cmark & 0.95 & 0.52 & 0.41 & 0.20 \\
\cmark & \cmark & \xmark & \cmark & \cmark & 0.79 & 0.37 & 0.27 & 0.13 \\
\cmark & \cmark & \cmark & \xmark & \cmark & 0.65 & 0.33 & 0.24 & 0.12 \\
\cmark & \cmark & \cmark & \cmark & \xmark & 0.88 & 0.38 & 0.29 & 0.14 \\
\bottomrule
\end{tabular}

\end{table}

\noindent\textbf{Effect of Trajectory Preference Optimization}
\label{sec:tpo-effect}
We assess the contribution of the second stage by comparing the best \emph{SFT-only} checkpoint against the model after \emph{Trajectory Preference Optimization (TPO)}—our DPO instantiation in Sec.~\ref{sec:dpo} that prefers trajectories with lower $\ell_2$ displacement. The results are reported as averages of the metrics under each protocol over a 3\,s horizon.

\begin{table}[t]
\centering
\setlength{\tabcolsep}{4pt}
\renewcommand{\arraystretch}{1.05}
\footnotesize
\caption{Impact of TPO on NuScenes. We report average $\ell_2$ (m, $\downarrow$) and collision rate (\%, $\downarrow$) under both \textsc{ST-P3} and \textsc{UniAD} protocols.}
\label{tab:tpo}
\begin{tabular}{l|cc|cc}
\toprule
\multirow{2}{*}{\textbf{Stage}} & \multicolumn{2}{c|}{\textbf{$\ell_2$ (m) $\downarrow$}} & \multicolumn{2}{c}{\textbf{Collision (\%) $\downarrow$}} \\
 & ST\text{-}P3 & UniAD & ST\text{-}P3 & UniAD \\
\midrule
SFT        & 0.39 & 0.78 & 0.11 & 0.23 \\
SFT + TPO  & \textbf{0.31} & \textbf{0.61} & \textbf{0.10} & \textbf{0.22} \\
\bottomrule
\end{tabular}

\end{table}

TPO yields a substantial reduction in average $\ell_2$ displacement across both protocols, confirming that preference learning driven by trajectory quality sharpens planning accuracy. Though post-SFT collision rate is already low (e.g., $0.11$ under \textsc{ST-P3}), TPO still further reduces the collision rate by introducing more accurate trajectory planning.

\noindent\textbf{Compatibility with Different VLM Backbones}
\label{sec:backbone-compat}
We further test our training recipe on alternative backbones to assess portability. Applying the same SFT\,$\rightarrow$\,TPO pipeline to Qwen2.5-VL-7B and InternVL-3.5-8B yields consistently strong results, demonstrating that the procedure is \emph{backbone-agnostic} and can convert a range of VLMs into trajectory-planning agents.

\begin{table}[t]
\centering
\setlength{\tabcolsep}{3.5pt}  
\renewcommand{\arraystretch}{1.05}
\footnotesize
\caption{Backbone compatibility of \textbf{LLaViDA}. S = ST\text{-}P3, U = UniAD.}
\label{tab:backbone}
\begin{tabular}{@{}p{.58\columnwidth}cc@{}}
\toprule
\textbf{Backbone} & \textbf{$L_2$ (S/U) $\downarrow$} & \textbf{CR (S/U) $\downarrow$} \\
\midrule
LLaVA\textendash NeXT\textendash LLaMA\textendash 3\textendash 8B & \textbf{0.31}/\textbf{0.61} & \textbf{0.10}/\textbf{0.22} \\
Qwen2.5\textendash VL\textendash 7B                               & 0.31/0.62 & 0.10/0.22 \\
InternVL\textendash 3.5\textendash 8B                              & 0.32/0.63 & 0.11/0.23 \\
\bottomrule
\end{tabular}

\end{table}

\begin{figure}[!htbp]
    \centering
    \includegraphics[width=\linewidth]{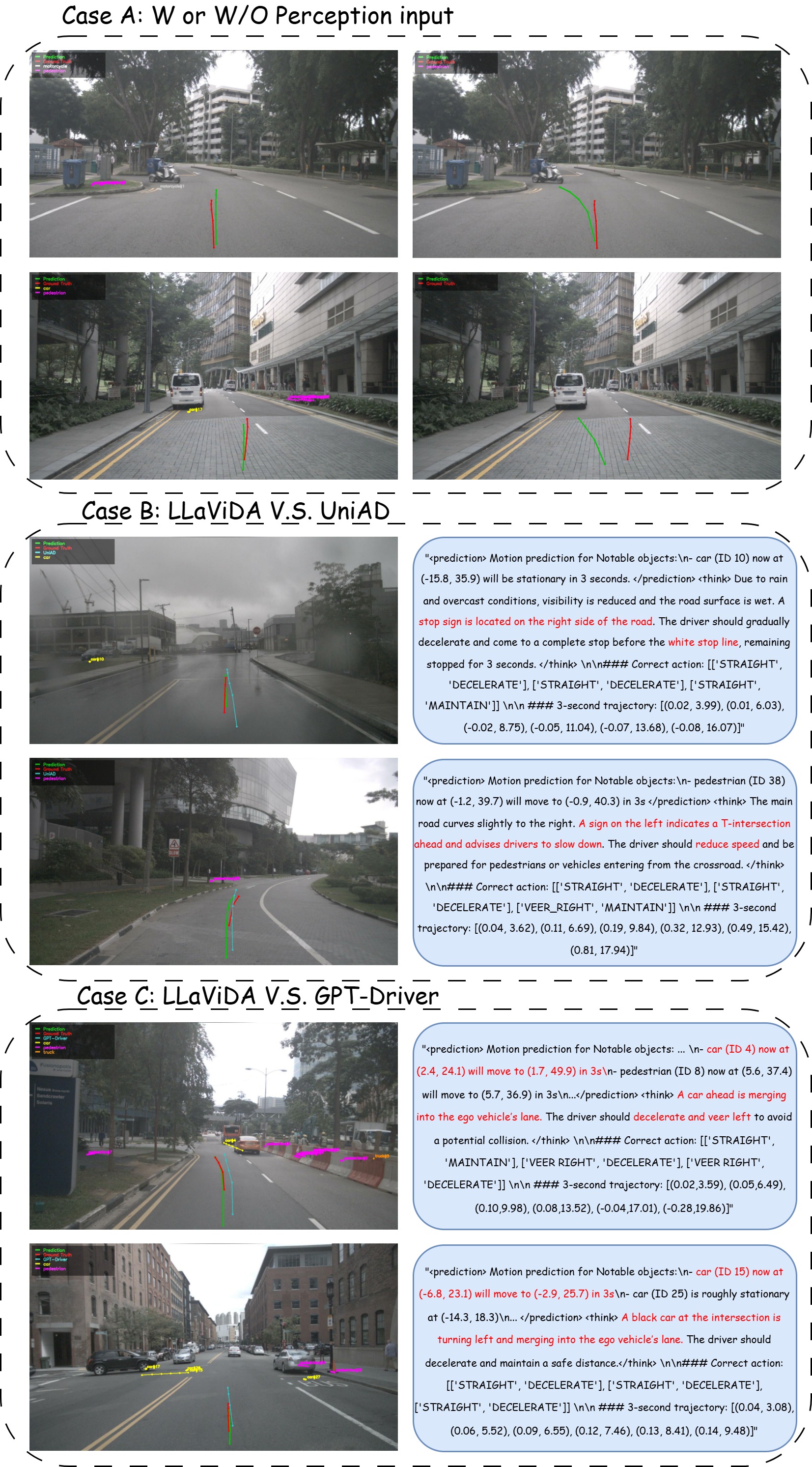}
    \caption{Representative qualitative results. All trajectories are overlaid on the same front-view camera images: red indicates the ground-truth trajectory, green represents the prediction from our method, and cyan denotes the baseline prediction. Text boxes contain the corresponding textual output from our pipeline.}

    \label{fig:case}
    
\end{figure}

\subsection{Efficiency Optimization}
\label{sec:efficiency}

Explicit reasoning improves interpretability and robustness, but long token roll-outs become the main bottleneck for real-time use. We therefore optimize \textbf{LLaViDA} along three axes. \textbf{(i) Direct-output decoding.} During SFT, we mix samples with full reasoning+motion labels and samples with only meta-action+trajectory labels. This teaches the model implicit reasoning and multi-object motion prediction, while also enabling it to \emph{directly} emit meta actions and trajectories at inference, cutting roll-out length. \textbf{(ii) View reduction.} While six surround views offer rich context, they inflate vision tokens even with $2\times 2$ pooling. Following prior observations that the front view dominates planning accuracy ~\cite{tian2024drivevlm}, we switch to front-only images in both training and inference, lowering prefill cost with a small accuracy drop. \textbf{(iii) KV caching.} Since the system prompt and schema are constant across steps, we cache key–value states to reduce repeated prefill in successive planning cycles.

\begin{table*}[t]
\centering
\setlength{\tabcolsep}{2pt}
\renewcommand{\arraystretch}{1.05}
\scriptsize
\caption{Performance and latency trade-offs. Latency averaged over the NuScenes test split (6019 samples) on an NVIDIA A100 GPU.}
\label{tab:efficiency}
\begin{tabular}{@{}p{.32\textwidth}ccccc@{}}
\toprule
\textbf{Version} &
\textbf{L2 (ST-P3/UniAD) $\downarrow$} &
\textbf{CR (ST-P3/UniAD) $\downarrow$} &
\textbf{Prefill (ms)} &
\textbf{Roll-out (ms)} &
\textbf{Total (ms)} \\
\midrule
Full (six views, with reasoning)   & \textbf{0.31 / 0.61} & \textbf{0.10 / 0.22} & 643 & 1780 & 2423 \\
+ Direct output        & 0.34 / 0.66          & 0.12 / 0.25          & 643 & 535  & 1178 \\
+ Front camera only                  & 0.35 / 0.68          & 0.13 / 0.26          & 362 & 535  & 897  \\
+ KV cache       & 0.35 / 0.68          & 0.13 / 0.26          & 239 & 535  & 776  \\
\bottomrule
\end{tabular}
\vspace{-2mm}
\end{table*}

\noindent\textbf{Discussion.} Mixing direct-output supervision enables large roll-out savings with minimal accuracy loss; pruning to the front view and enabling KV cache further reduce prefilling time. The final configuration achieves substantially lower end-to-end latency while maintaining competitive $L_2$/collision metrics, making LLaViDA more suitable for real-time deployment.

\subsection{Case Study}
\label{sec:qualitative}

We illustrate three representative scenarios in Figure~\ref{fig:case}, highlighting why \textbf{LLaViDA} achieves strong planning quality: (A) the effect of adding structured perception input, (B) semantic scene understanding compared with a modular non-autoregressive planner (UniAD), and (C) the impact of explicit motion prediction compared with an autoregressive baseline (GPT-Driver). In all panels, the predicted ego trajectory is overlaid on the front camera view; ground truth is shown in red for reference.

\textbf{Case A: with vs. without perception input.}
Figure~\ref{fig:case} A shows that supplying key-object states (BEV location, velocity, class) materially improves planning. In the top example, the perception-free variant fails to localize a motorcycle on the left and produces a trajectory that clips its path; with perception, LLaViDA identifies the motorcycle and yields. In the bottom example, the perception-free variant misses a van parked along the left side, again leading to a risky plan; with perception, the model steers a safe buffer. Prior work has noted that VLMs can struggle with precise spatial localization and may hallucinate small or occluded objects~\cite{tian2024drivevlm}; injecting a lightweight 3D perception module provides reliable anchors and reduces these errors.

\textbf{Case B: LLaViDA vs. UniAD (semantic grounding).}
Figure~\ref{fig:case} B contrasts LLaViDA with UniAD on scenes that require reading signs and markings. In the top scene, LLaViDA recognizes a stop sign and the painted stop line, reasons to decelerate, and halts before the line; UniAD continues at near-constant speed. In the bottom scene, LLaViDA identifies a "SLOW" warning and reduces speed accordingly, while UniAD again maintains speed. These cases illustrate the advantage of language-grounded reasoning for compliance with traffic semantics.

\textbf{Case C: LLaViDA vs. GPT-Driver (motion prediction).}
Figure~\ref{fig:case} C demonstrates the benefit of explicit multi-agent motion prediction. LLaViDA infers that vehicles ahead are merging into the ego lane, reasons about the narrowing gap, and plans a protective slowdown and offset. GPT-Driver, which conditions primarily on current object locations without forecasting, fails to anticipate the merge and emits a riskier plan. The examples underline that forecasting agent intentions is critical for safe short-horizon planning.

\vspace{-0.6em}

\section{Conclusion}
\label{sec:conclusion}

We introduced \textbf{LLaViDA}, a vision–language model expert for autonomous driving trajectory planning that unifies perception, reasoning, and motion prediction into a single interpretable process. Through a two-stage pipeline—supervised fine-tuning on the curated \emph{NuScenes-TP} dataset followed by trajectory preference optimization—LLaViDA efficiently adapts general-purpose VLMs into accurate, reasoning-driven planners. Experiments on NuScenes open-loop performance achieve both interpretability and real-time efficiency after inference optimizations. Due to computational resource constraints, we have not yet scaled training to larger datasets such as nuPlan, which we leave for future work to further enhance robustness and generalization.

{\small
\bibliographystyle{ieeenat_fullname}
\bibliography{11_references}
}

\ifarxiv \clearpage \appendix \section{Details for \textit{NuScenes-TP} Construction}
\label{app:nuscenes-tp}

We use NuScenes~\cite{caesar2020nuscenes}, a state-of-the-art dataset for autonomous driving composed of multiple modalities of sensor data and control actions. Since no meta-actions are present in the native dataset, we design a novel protocol to construct a dataset suitable for training language models in autonomy. We first present logic that ingests driving control history and generates meta-action descriptions in natural language. We then describe our procedure for using prompts to cue GPT reasoning. 

\subsection{Meta-Action Definition and Labeling}
\label{app:meta}

\textbf{Horizon and sampling.}
Because the framework predicts a 3\,s future ego trajectory, we define \emph{3\,s meta-actions} aligned to this horizon. Two complementary formulations are used: (i) \emph{local per-interval} actions over \([0{\to}1]\), \([1{\to}2]\), \([2{\to}3]\) seconds (used as training labels), which correlate tightly with per-second waypoints; and (ii) \emph{cumulative} actions over \([0{\to}1]\), \([0{\to}2]\), \([0{\to}3]\) seconds (used to verify GPT-synthesized reasoning). Waypoints are sampled at 2\,Hz, while meta-actions are sampled at 1\,Hz since 0.5\,s deltas are too subtle for reliable action discrimination.

\textbf{Annotation seed and two labelers.}
We randomly sample 1{,}000 training samples from NuScenes and obtain human expert labels for meta-actions. We then instantiate two automatic labelers: (a) a \emph{rule-based} labeler driven by per-second yaw and speed deltas; and (b) a lightweight \emph{model-based} classifier (a few Transformer blocks plus a classification head) trained on the 1k expert set.

\textbf{Rule-based labeler.}
The rule-based labeler takes in a history of control signals and outputs the meta-action in natural language that the agent took. We factor meta-actions into lateral and longitudinal components to represent the orthogonal components of steering (lateral) and throttle/brake (longitudinal). Let $\Delta\!\psi$ be the absolute yaw change (degrees) over 1\,s and $\Delta\!v$ the speed change (m/s) over 1\,s.
\emph{Lateral:} \(\Delta\!\psi < 5^\circ \Rightarrow\) keep; \(5^\circ \le \Delta\!\psi < 20^\circ \Rightarrow\) veer (L/R by sign); \(\Delta\!\psi \ge 20^\circ \Rightarrow\) turn (L/R by sign).
\emph{Longitudinal:} \(\Delta\!v \ge +0.25\,\mathrm{m/s} \Rightarrow\) accelerate; \(\Delta\!v \le -0.25\,\mathrm{m/s} \Rightarrow\) decelerate; a sustained decrease \(\Delta\!v \le -0.5\,\mathrm{m/s}\) until \(v < \varepsilon\) (e.g., \(0.1\,\mathrm{m/s}\)) is labeled brake-to-stop.
The joint meta-action is the Cartesian product of lateral and longitudinal decisions (details of mapping and tie-breaking are provided in the code release). For training labels we use the \emph{local per-interval} actions; for reasoning verification we accept GPT outputs if any \emph{cumulative} action \([0{\to}t]\) matches (\(t\!\in\!\{1,2,3\}\)).

\textbf{Empirical choice.}
Table~\ref{tab:meta-rule-vs-model} compares rule- vs.\ model-based meta-action labels when used in SFT. Rule-based labeling yields slightly better planning metrics for our pipeline and is therefore adopted as default.

\begin{table}[t]
\centering
\setlength{\tabcolsep}{3pt}
\renewcommand{\arraystretch}{1.05}
\scriptsize
\caption{Meta-action labeler comparison (3\,s horizon). S = ST\text{-}P3, U = UniAD.}
\label{tab:meta-rule-vs-model}
\begin{tabular}{@{}lcc@{}}
\toprule
\textbf{Meta-action labeler} & \textbf{$L_2$ (S/U) $\downarrow$} & \textbf{CR (S/U) $\downarrow$} \\
\midrule
Rule-based   & \textbf{0.31} / \textbf{0.61} & \textbf{0.10} / \textbf{0.22} \\
Model-based  & 0.33 / 0.63 & 0.11 / 0.23 \\
\bottomrule
\end{tabular}
\end{table}

\textbf{Rationale for two meta-action formulations.}
Training uses local per-interval labels because they align cleanly with per-second waypoint targets, improving supervision of short-horizon maneuvers. Reasoning verification uses cumulative labels to accommodate the flexible temporal abstractions of GPT: a correct forecast at any \([0{\to}t]\) checkpoint constitutes a valid explanation even if the precise second-by-second decomposition differs.
\subsection{Prompts for GPT Reasoning Generation and Verification}
\label{app:prompts}

We provide prompt for the reasoning generation and verification below for reproducibility:

\begin{lstlisting}[language=Python,basicstyle=\ttfamily\footnotesize,breaklines=true]
def build_autonomous_driving_prompt(
        ego_state,
        camera_info_dict: dict,
        objects_description: str = None,
        use_base64: bool = True):
    """
    Construct messages for a ChatCompletion-style API that contain:
      - system role
      - task instructions + one demo
      - six camera views (image or path text)
      - optional key-object summary paragraph
    """

    # ---------- system message ---------------------------------------------
    system_prompt = (
        "You are an autonomous-driving vision analyst.\n"
        "Think step-by-step like an experienced human driver observing the surroundings. "
        "Output ONLY the three numbered sections below. Do NOT prescribe steering or speed commands."
    )

    # ---------- user instructions & demo -----------------------------------
    user_prompt = """\
### Task
From the inputs (six surround-view images, ego state, and a key-object summary),
produce a concise situation report with three numbered sections:

1) Potential effects - effects caused by the positions and movements of notable objects.
2) Road & Contextual Factors - lane geometry, surface condition, visibility, occlusions,
   traffic signs, traffic lights, road signs, etc.
3) Situation Snapshot - what the driver's "mental picture" looks like now and possible driving plans.

### Few-shot Example

Camera Views (sample):
- front-left: parked cars at curb
- front: blue sedan 30 m ahead, braking
- front-right: clear sidewalk
- back-left: black SUV closing in left lane
- back: clear
- back-right: cyclist 20 m behind

Key-Object Summary (sample):
All coordinates are given in a 2-D egocentric plane...
Notable objects:
 1. A car currently at (3.2, 30.0) and expected to move toward (3.1, 28.0).
 2. A cyclist currently at (-1.5, -20.0) and expected to remain roughly stationary.

Model Output (human-style chain of thought)
1) Potential effects
   The SUV is gaining on the sedan and may cut into my lane to avoid slowing.
   If the SUV merges, my forward gap shrinks; leave room to brake or change lanes.
   The cyclist poses minimal immediate risk but occupies a potential escape route to the right.
   Parked cars on the front-left restrict lateral escape; mirrors show no vehicles in the blind spot.

2) Road & Contextual Factors
   Dry pavement, lane markings clear; no work zones or debris visible.
   Morning sun low on horizon may cause glare for oncoming traffic.
   Road remains straight for ~200 m; an overpass ahead may create a brief shadowed section.

3) Situation Snapshot
   I am following a blue sedan that is braking; the left lane has a black SUV closing quickly;
   the right-rear cyclist remains steady; the road ahead appears straight and clear.

---
Analyse the new scene below:
""".lstrip()

    # ---------- assemble messages -----------------------------------------
    system_msg = {"role": "system", "content": system_prompt}
    user_msgs  = [{"type": "text", "text": user_prompt}]

    # (a) camera views
    user_msgs.append({"type": "text", "text": "Camera Views:"})
    if use_base64:
        for view, path in camera_info_dict.items():
            user_msgs.append({"type": "text", "text": f"{view}:"})
            img_b64 = encode_image(path)  # assumes helper exists
            user_msgs.append({
                "type": "image_url",
                "image_url": {"url": f"data:image/jpeg;base64,{img_b64}"}
            })
    else:
        for view, path in camera_info_dict.items():
            user_msgs.append({"type": "text", "text": f"{view}: {path}"})

    # (b) ego + key-object summary (if any)
    user_msgs.append({"type": "text", "text": "\nEgo state:"})
    user_msgs.append({"type": "text", "text": ego_state})
    if objects_description:
        user_msgs.append({"type": "text", "text": "\nKey-Object Summary:"})
        user_msgs.append({"type": "text", "text": objects_description})

    # assistant kick-off token
    assistant_msg = {"role": "assistant", "content": "Step-by-step reasoning:"}

    return system_msg, {"role": "user", "content": user_msgs}, assistant_msg
\end{lstlisting}

\begin{lstlisting}[language=Python,basicstyle=\ttfamily\footnotesize,breaklines=true]
def build_verify_prompt(ego_state, description, image_path, reasoning_context: str,
                        speed: float, add_image=True):
    """
    Construct a concise prompt for an LLM that returns a driving
    meta-action pair and a confidence score in [0, 5].
    """
    system_prompt = """
You are an autonomous-driving assistant.
Input: key-object description + reasoning context + ego state and speed + camera images.
Task: decide what the ego vehicle should do from the lateral and longitudinal aspects.

Output format (no extra text):
(['<LATERAL>', '<LONGITUDINAL>'], <CONFIDENCE>)    # confidence in [0, 5]

Allowed meta-actions
- Lateral: VEER_LEFT | VEER_RIGHT | STRAIGHT | TURN_LEFT | TURN_RIGHT
- Longitudinal: ACCELERATE | MAINTAIN | DECELERATE | BRAKE_TO_STOP

Decision rules
1. Avoid collisions with other objects; keep safe gaps.
2. Stay on drivable surface.
3. Keep reasonable speed when the road is clear.
4. Turn at low speed while decelerating.

Considerations (IMPORTANT)
- Lateral:
  a) Check roadway geometry first. If the main lane curves ahead, select the action that
     follows the curve (never output STRAIGHT in this case).
  b) Then account for pedestrians, vehicles, or other obstacles and steer to avoid
     any potential collision.
- Longitudinal:
  a) Begin with the current speed.
  b) Decide on a change:
     - If the vehicle is moving too slowly for conditions, ACCELERATE.
     - If it is too fast or needs extra margin, DECELERATE.
     - Otherwise, MAINTAIN the present speed.
""".strip()

    # Assemble chat messages
    system_content = [{"type": "text", "text": system_prompt}]
    user_content = [
        {"type": "text", "text": f"Key object description:\n{description}"},
        {"type": "text", "text": f"Reasoning context:\n{reasoning_context}"},
        {"type": "text", "text": f"Ego state: {ego_state}"},
        {"type": "text", "text": f"Ego speed: {speed} m/s"},
    ]
    if add_image:
        user_content.append({"type": "text", "text": "Camera Views:"})
        for view, path in image_path.items():
            user_content.append({"type": "text", "text": f"{view}:"})
            encoded = encode_image(path)  # assumes helper exists
            user_content.append({
                "type": "image_url",
                "image_url": {"url": f"data:image/jpeg;base64,{encoded}"}
            })

    assistant_content = [{"type": "text", "text": "Meta-action and confidence:"}]
    return system_content, user_content, assistant_content
\end{lstlisting}

\subsection{Prompt Construction for Training Samples}
\label{app:prompt-train}

 Each training sample uses a paired prompt–completion. The \texttt{human\_message\_value} asks the VLM to (i) build a concise context with short-horizon motion prediction, (ii) write step-by-step reasoning, (iii) output a strictly formatted 3 s meta-action sequence (1 Hz), and (iv) output a strictly formatted 3 s trajectory with 6 waypoints (2 Hz). The completion \texttt{gpt\_message\_value} carries the target \texttt{<prediction>} (object forecasts), \texttt{<think>} (reasoning), the canonical meta-actions, and the 3 s trajectory.

\lstset{basicstyle=\ttfamily\footnotesize,breaklines=true,columns=fullflexible,keepspaces=true}
\begin{lstlisting}[language=Python]
human_message_value = (
    "You are provided with six synchronized camera images captured from the ego-vehicle "
    "in the following order: rear, rear-left, rear-right, front, front-left, and front-right. "
    f"The current state information of the ego-vehicle is: {ego}. "
    f"The current perceived notable objects are listed here: {perception}. "
    "<task> First, formulate a concise context that integrates scene perception and short-term motion prediction. "
    "You should provide approximate 2-D Bird-Eye-View coordinates for every notable object's future waypoints in 3 seconds "
    "in your reasoning process. The higher the ego velocity is, the more distant objects you should consider. "
    "Then, based on perception and prediction, provide your chain-of-thought reasoning about the current driving scene, "
    "integrating potential effects of the notable objects, road and contextual factors, semantic grounding, "
    "and the driver's mental picture. "
    "After that, derive an appropriate driving decision sequence for 3 seconds ahead (one decision per second) and return it exactly "
    "as a list of lists in the format [['<LATERAL>', '<LONGITUDINAL>'], ['<LATERAL>', '<LONGITUDINAL>'], ['<LATERAL>', '<LONGITUDINAL>']]. "
    "Finally, based on all context and the derived driving decisions, plan a safe, feasible 3-second trajectory of 6 waypoints and return it exactly "
    "as a list of waypoint tuples in the format [(x1,y1), (x2,y2), (x3,y3), (x4,y4), (x5,y5), (x6,y6)] "
    "(one waypoint per 0.5 s). </task> "
    "<meta action pool> Permissible lateral actions: VEER_LEFT | VEER_RIGHT | CHANGE_LANE_LEFT | CHANGE_LANE_RIGHT | STRAIGHT | TURN_LEFT | TURN_RIGHT. "
    "Permissible longitudinal actions: ACCELERATE | MAINTAIN | DECELERATE | BRAKE_TO_STOP. </meta action pool> "
    "<coordinate instruction> Coordinates: X-axis is lateral (left/right), Y-axis is longitudinal (forward). "
    "You are at (0,0). Units: meters. </coordinate instruction>"
)

gpt_message_value = (
    f"<prediction> {prediction} </prediction> "
    f"<think> {reasoning_text} </think>\n\n"
    f"### Correct action: {complete_action}\n\n"
    f"### 3-second trajectory: {trajectory}"
)
\end{lstlisting}
\paragraph{Variable explanations.}
\begin{itemize}
  \item \texttt{ego}: Ego vehicle state string (speed, 2 s history of waypoints, heading/yaw).
  \item \texttt{perception}: String of key-object states (per-object BEV location and velocity).
  \item \texttt{prediction}: Structured object waypoint forecasts over 3 s (used in completion).
  \item \texttt{reasoning\_text}: Natural language chain of thought (scene semantics, weather, road layout, agent intents).
  \item \texttt{complete\_action}: Three meta-actions, one per second, formatted as \texttt{[['<LATERAL>', '<LONGITUDINAL>'], ...]}.
  \item \texttt{trajectory}: Six ego waypoints \texttt{[(x1,y1),...,(x6,y6)]} at 0.5 s intervals.
\end{itemize}

\section{Experiment details}
\subsection{Training Hyperparameters}
\label{app:hyperparams}

We provide the core training hyperparameters for reproducibility. All training and evaluation are implemented on 4 A100 GPUs.

\begin{table}[t]
\centering
\setlength{\tabcolsep}{4pt}
\renewcommand{\arraystretch}{1.05}
\scriptsize
\caption{Core training hyperparameters.}
\label{tab:hparams}
\begin{tabular}{@{}lcc@{}}
\toprule
\textbf{Hyperparameter} & \textbf{SFT} & \textbf{TPO} \\
\midrule
Per-device batch size & 1 & 1 \\
Number of GPUs & 4 & 4 \\
LLM learning rate & \(5\times10^{-7}\) & \(5\times10^{-7}\) \\
Vision encoder learning rate & \(5\times10^{-8}\) & frozen \\
Epochs & 3 & 1 \\
\bottomrule
\end{tabular}
\end{table}

\subsection{Metric Calculation Details (ST\text{-}P3 vs.\ UniAD)}
\label{app:metrics}

\paragraph{Setup and notation.}
The predicted and ground-truth ego trajectories over a 3 s horizon (sampled at 2 Hz) are
\(
\hat{T}=\{(\hat{w_1}_t,\hat{w_2}_t)\}_{t=1}^{H},\;
T^{\ast}=\{({w_1}^{\ast}_t,{w_2}^{\ast}_t)\}_{t=1}^{H}
\)
with \(H=6\). A per-timestep visibility mask \(m_t\in\{0,1\}\) (from \texttt{gt\_traj\_mask}) down-weights invalid steps. The occupancy maps \(\{S_t\}_{t=1}^{H}\in\{0,1\}^{200\times 200}\) use a BEV grid that covers \([-50,50]\) m by \([-50,50]\) m at 0.5 m resolution. The ego footprint is an axis-aligned rectangle of length \(4.084\,\mathrm{m}\) and width \(1.85\,\mathrm{m}\) placed at the trajectory center at each \(t\) (orientation/yaw is not applied).

\subsection*{L2 displacement}
Per-timestep Euclidean error:
\[
d_t \;=\; \big\|(\hat{x}_t,\hat{y}_t)-(x^{\ast}_t,y^{\ast}_t)\big\|_2.
\]
Aggregation differs by protocol:
\begin{itemize}
\item \textbf{ST\text{-}P3:} for \(k\in\{1,2,3\}\), report the mean up to the horizon,
\(
\mathrm{L2}@k\mathrm{s}=\frac{1}{2k}\sum_{t=1}^{2k} d_t.
\)
\item \textbf{UniAD:} for \(k\in\{1,2,3\}\), report the single-step error at \(t=2k\),
\(
\mathrm{L2}@k\mathrm{s}=d_{2k}
\)
(indices \(t=2,4,6\)).
\end{itemize}

\subsection*{Collision rate}
At each step \(t\), rasterize the fixed-size ego rectangle centered at \((\hat{x}_t,\hat{y}_t)\) into a pixel set \(\mathcal{B}_t\) in BEV. Define box collision
\[
c^{\mathrm{box}}_t \;=\; \mathbb{1}\ \left[\;\exists (r,c)\in\mathcal{B}_t\ \text{s.t.}\ S_t[r,c]=1\;\right].
\]
Exclude steps where the \emph{ground-truth} box already collides:
\(
\tilde{c}_t = c^{\mathrm{box}}_t \cdot (1-c^{\mathrm{box,GT}}_t).
\)
Aggregation mirrors L2:
\begin{itemize}
\item \textbf{ST\text{-}P3:} mean up to horizon,
\(
\mathrm{CR}@k\mathrm{s}=\frac{1}{2k}\sum_{t=1}^{2k}\tilde{c}_t.
\)
\item \textbf{UniAD:} value at horizon,
\(
\mathrm{CR}@k\mathrm{s}=\tilde{c}_{2k}.
\)
\end{itemize}

\subsection*{Coordinate handling and implementation notes}
\begin{itemize}
\item \textbf{BEV grid.} Resolution \(\mathrm{dx}=[0.5,0.5]\) m and start offsets \(\mathrm{bx}=[-50+0.25,-50+0.25]\) m give a \(200\times 200\) map.
\item \textbf{Axis alignment.} The ego box is axis-aligned (no yaw) for collision checks.
\item \textbf{Protocol-specific flips.} For ST\text{-}P3, occupancy maps are flipped on both spatial axes at load time; trajectories are flipped once on the \(x\)-axis inside the evaluator. For UniAD, trajectories are flipped on \(x\) in \texttt{update} and again inside \texttt{evaluate\_coll} (net zero), and occupancy maps are used as-is. These choices reproduce the public codebase.
\end{itemize}

\subsection*{Summary of protocol differences}
\begin{table}[t]
\centering
\setlength{\tabcolsep}{3pt}
\renewcommand{\arraystretch}{1.05}
\scriptsize
\begin{tabular}{@{}p{.45\columnwidth}p{.23\columnwidth}p{.23\columnwidth}@{}}
\toprule
 & \textbf{ST\text{-}P3} & \textbf{UniAD} \\
\midrule
Horizon aggregation & Mean over steps \(1..2k\) & Single step at \(t=2k\) \\
L2 @ \(k\) s & \(\tfrac{1}{2k}\sum_{t\le 2k} d_t\) & \(d_{2k}\) \\
Collision @ \(k\) s & \(\tfrac{1}{2k}\sum_{t\le 2k}\tilde{c}_t\) & \(\tilde{c}_{2k}\) \\
Occ map prep & Flip both axes & No flip \\
Traj \(x\)-flip & Once in evaluator & Twice (net zero) \\
Ego footprint & \(4.084\,\mathrm{m}\times 1.85\,\mathrm{m}\), axis-aligned & Same \\
Rate & 2 Hz (6 steps in 3 s) & 2 Hz (6 steps in 3 s) \\
\bottomrule
\end{tabular}
\end{table}

\noindent
In short, ST\text{-}P3 emphasizes average performance across the horizon, whereas UniAD evaluates accuracy exactly at the 1/2/3 s horizons. Collision is computed on the rasterized ego \emph{box} (not only the center point) and excludes steps where the ground-truth box collides. Please check the original paper for more metric details.

\subsection{Evaluation of Meta-Action Prediction}
\label{app:meta-eval}

Meta actions bridge semantic reasoning and numeric trajectory generation. To verify that the final model learns consistent mappings from context to action, we evaluate \emph{lateral} and \emph{longitudinal} decisions in two regimes over a 3 s horizon: (i) \textbf{Per-interval} correctness at 1 s, 2 s, 3 s; and (ii) \textbf{Cumulative} correctness, which counts success at time $k$ only if all actions up to $k$ are correct.

\begin{table}[t]
\centering
\setlength{\tabcolsep}{3pt}
\renewcommand{\arraystretch}{1.05}
\scriptsize
\caption{Meta-action accuracy (\%). Per-interval vs.\ cumulative correctness at 1 s, 2 s, 3 s.}
\label{tab:meta-accuracy}
\begin{tabular}{lccc|ccc}
\toprule
\multirow{2}{*}{Type} & \multicolumn{3}{c}{Per-interval (\%)} & \multicolumn{3}{c}{Cumulative (\%)} \\
\cmidrule(lr){2-4}\cmidrule(lr){5-7}
& 1 s & 2 s & 3 s & 1 s & 2 s & 3 s \\
\midrule
Lateral       & 89.3 & 81.2 & 64.7 & 89.3 & 71.1 & 39.9 \\
Longitudinal  & 91.1 & 83.5 & 68.3 & 91.1 & 75.4 & 48.7 \\
\bottomrule
\end{tabular}
\end{table}

\paragraph{Summary.}
The model attains strong 1 s accuracy (lateral 89.3\%, longitudinal 91.1\%), and remains robust at 2 s. Accuracy declines by 3 s, with a larger drop under \emph{cumulative} evaluation (lateral 39.9\%, longitudinal 48.7\%) due to error compounding across steps. Longitudinal decisions are consistently a few points higher than lateral, suggesting the model more reliably regulates speed than precise lateral manoeuvres over longer horizons. These results support our design where meta actions serve as an effective bridge from textual reasoning to precise trajectory generation while highlighting scope for improving long-horizon consistency.

\section{Visualization}

We randomly sampled and visualized more outputs from our best checkpoint in Figure~\ref{fig:vis}.
\begin{figure*}[!htbp]
    \centering
    \includegraphics[width=\linewidth]{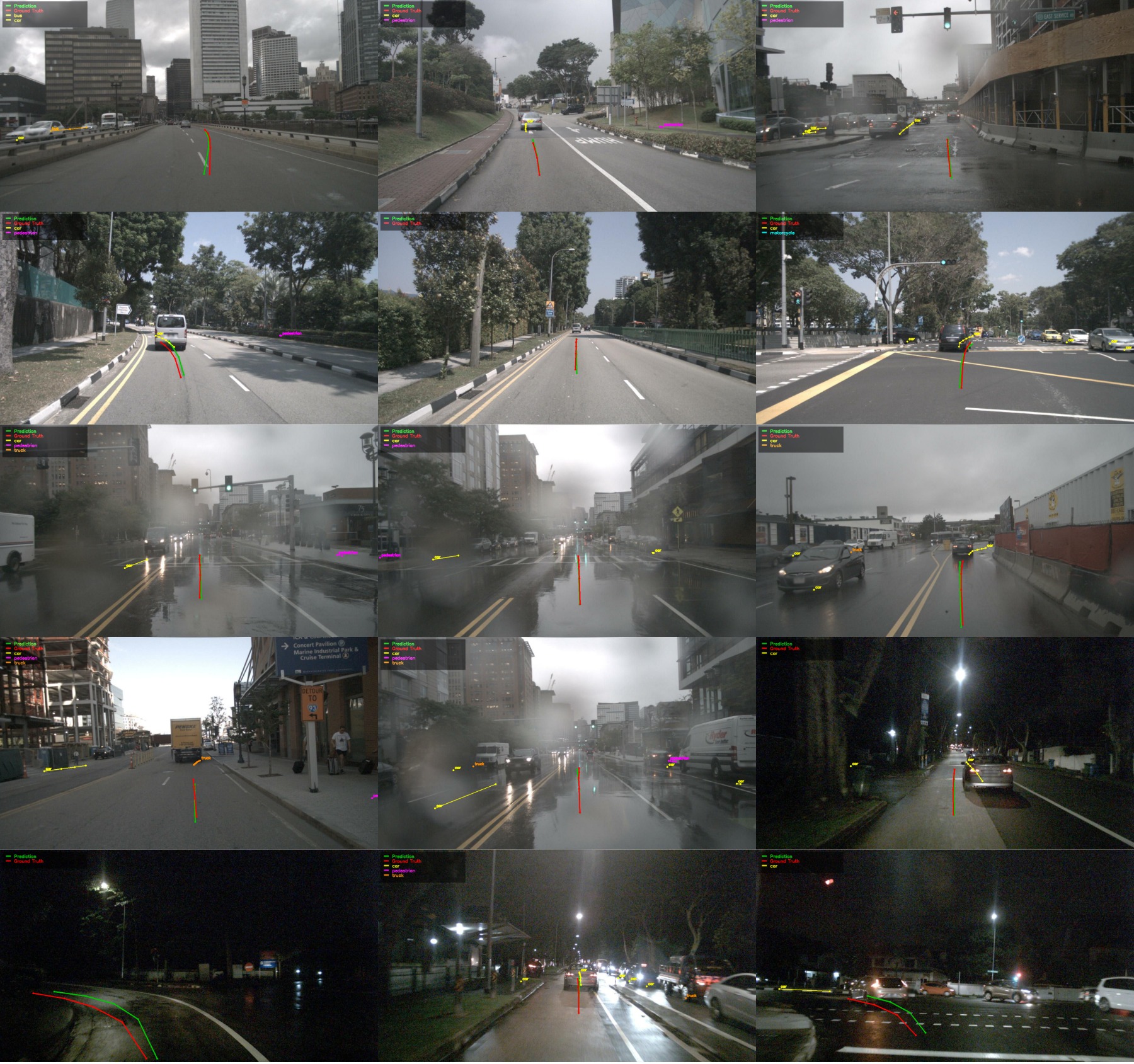}
    \caption{Visualization sampled from NuScenes test split. Ground truth trajectory in red and predicted trajectory in green.}

    \label{fig:vis}
    
\end{figure*} \fi

\end{document}